\documentclass[runningheads]{llncs}

\usepackage[T1]{fontenc}
\usepackage{graphicx}

\usepackage{amsmath}
\usepackage{amssymb}
\usepackage{color}
\usepackage{xspace}
\usepackage{booktabs}
\usepackage{multirow}
\usepackage{multicol}
\usepackage{enumitem}

\newcommand{\owls}{OWL-S\xspace}
\newcommand{\wsmo}{WSMO\xspace}
\newcommand{\kg}{KG\xspace}
\newcommand{\kgs}{KGs\xspace}
\newcommand{\iope}{IOPE\xspace}
\newcommand{\iopes}{IOPEs\xspace}

\newcommand{\psm}{PSM\xspace}
\newcommand{\psms}{PSMs\xspace}
\newcommand{\DimE}{\mathcal{E}}
\newcommand{\DimD}{\mathcal{D}}
\newcommand{\DimG}{\mathcal{G}}
\newcommand{\DimR}{\mathcal{R}}

\newcommand{\AAP}{\textsc{AAP}\xspace}
\newcommand{\AAPs}{\textsc{AAP}s\xspace}

\begin{document}

\title{Discoverable Agent Knowledge — A Formal Framework for Agentic KG Affordances (Extended Version)\thanks{A shorter version of this paper, including the worked example, was presented at EKAW 2026. 
This extended version provides a fuller
formal treatment of the Agentic Discoverability dimension (including a
set-theoretic characterisation of metadata-assessable fitness) and a
more detailed account of the relationship between the \AAP
framework and the ontological continuum~\cite{daga2026continuum}.}}

\titlerunning{A Formal Framework for Agentic KG Affordances}

\author{Terry R. Payne\inst{1}\orcidID{0000-0002-0106-8731} \and
Valentina Tamma\inst{1}\orcidID{0000-0002-1320-610X} \and Enrico Daga\inst{2}\orcidID{0000-0002-3184-5407}}

\authorrunning{T.R. Payne et al.}

\institute{
School of Computer Science and Informatics, University of Liverpool, UK \email{\{T.R.Payne,V.Tamma\}@liverpool.ac.uk} \and
Open University: Milton Keynes, Milton Keynes, UK \email{enrico.daga@open.ac.uk}}

\maketitle

\begin{abstract}
Two decades ago, the Semantic Web Services community was asked how agents
with different ontological commitments could discover, compose, and invoke
web services coherently. The response was \owls and \wsmo: formally
grounded capability descriptions specifying what a service could do, what
the agent must already know for invocation to be epistemically sound, and
how ontological mismatches could be formally bridged.
Current Knowledge Graph (\kg) metadata standards such as \textsc{VoID} and \textsc{DCAT}
describe what a \kg contains, yet say nothing about what a specific
agent can prove from it, what closure assumptions govern empty
results, or whether the agent's task vocabulary is grounded in the schema.
Furthermore, in deployed \kgs, the governing schema Description Logic
(\textsc{DL}) and the operative entailment regime can diverge,
resulting in an epistemic failure mode that is invisible to current metadata.
We revisit and extend these insights for the \kg setting with a
four-dimensional framework: \emph{Semantic Expressivity},  \emph{Task-Relative Grounding}, \emph{Epistemic Trust Scope} and \emph{Agentic Discoverability}, together with one possible concrete formalisation.
Based on these four dimensions, we derive the \emph{Agentic Affordance Profile (\AAP)},
a semantic layer above \textsc{VoID} and \textsc{DCAT} enabling principled
\kg selection, composition, and failure diagnosis at agent planning time.
Together, they operationalise the affordance structure of the
Ontological Continuum~\cite{daga2026continuum} at the individual-agent
level.
A worked example, drawn from a variant of the OAEI Conference.owl ontology, concretely grounds the
framework and identifies the formal,
computational, and engineering work needed to realise \AAP-based
affordance matching at scale through a five-point research agenda.
\end{abstract}


\section{Introduction}
\label{sec:intro}
LLM-based agents that collaborate to solve composite tasks through the
discovery and integration of knowledge-based
resources~\cite{abouali2025agentic,wang2024survey} have driven a growing
interest in \emph{Agentic AI}. This raises open challenges around the
autonomy and governance of agents that reason over Knowledge Graphs
(\kgs)~\cite{calbimonte2023autonomy,kampik2022governance}.
Structured knowledge supports agent reasoning and decision
making~\cite{wooldridge1995intelligent}, constrains query interpretation
via ontological vocabularies, and provides factual content for agent
conclusions.
Yet \kgs increasingly exist as independently published, heterogeneous
resources that agents must discover, select, and evaluate dynamically,
rather than treating them as fixed, design-time commitments.
Furthermore, the means by which an agent \emph{assesses} whether a
given \kg is fit for a given task remains largely informal: it is either
pre-determined at design time, where the \kg, its schema assumptions,
and any required alignment or mediation are hard-coded as fixed system
dependencies, or conducted as an offline activity prior to deployment.
Standards such as \textsc{VoID}~\cite{alexander2011void} and
\textsc{DCAT}~\cite{maali2014dcat} describe what a \kg \emph{contains}
(triple counts, entity types, schema references, licensing); the
\textsc{SPARQL} Service
Description (\textsc{SPARQL-SD})~\cite{williams2013sparql-sd}
goes even further, declaring endpoint capabilities and, where specified,
supported entailment regimes~\cite{glimm2013sparqler}.
Yet none of these characterise a \kg's \emph{knowledge
affordance}~\cite{celino2026ka}: what task-relevant queries are
guaranteed to return meaningful results, what reasoning can be soundly
relied upon, or whether the vocabulary is intelligible to the agent
without mediation.
An LLM-orchestrated agent querying a \kg without such an assessment
risks unreliable retrieval and unsound conclusions from mismatched or incomplete content~\cite{chen2026datagents}.

The parallels with the Semantic Web Services community~\cite{ankolenkar2002damls,cabral2004approaches,WSMO2006,owls2004,martin2004owls,McIlraithSWS2001,wang2015survey,wang2012wsmo}
from two decades ago are striking; frameworks such as \owls~\cite{owls2004,martin2004owls} and
\wsmo~\cite{WSMO2006,wang2012wsmo} utilised ontologies for formally
describing how web services could be discovered, composed and queried.
Polleres et al.~\cite{Polleres26} characterise many of these
similarities, but also noted that current descriptions of agents are specified
in natural language (with input and output parameters appearing in
JSON Schema or via content types), and that such descriptions
\emph{``\ldots are prone to ambiguity as the semantics of the
functionality or capability is not precisely captured with natural
language descriptions alone\ldots''}.
A similar critique has also been echoed by researchers in the Multi-Agent Systems (MAS) community, who argue that many contemporary Agentic AI frameworks neglect decades of prior work on coordination, communication, interoperability and trust in distributed agent systems \cite{malfa2025large,wooldridge2024fetchai}.
What gave the Semantic Web Services frameworks much of their epistemic appeal was precisely this ontological grounding: in principle, agents operating under different ontological commitments could determine, prior to invocation, what could be soundly relied upon from a service and where mismatches would need to be formally bridged.

Given this convergence, we revisit the key structural insights from the
\owls/\wsmo tradition and extend them formally to the agentic \kg setting,
focusing on the \emph{inter-agent epistemic coherence
problem}~\cite{payne2014aamas,tamma2008savvy}: how a community of heterogeneous agents,
each grounded in its own ontology, can determine whether the knowledge
required for a given \kg interaction is coherent and meaningful.
We specify a four-dimensional formal framework for agentic \kg
affordance characterisation, and from it derive the
\emph{Agentic Affordance Profile (\AAP)} for \kgs: a semantic
layer above standards such as \textsc{VoID} and \textsc{DCAT}  enabling principled \kg
selection, composition, and failure diagnosis at agent planning time.
Essentially, an \emph{Agentic Affordance Profile} is a
task-relative semantic description of what an agent
can soundly retrieve, prove, and conclude from a given \kg, structured
along four dimensions:
\emph{Semantic Expressivity},  \emph{Task-Relative Grounding}, \emph{Epistemic Trust Scope} and \emph{Agentic Discoverability}.
We then present a research agenda identifying the formal, computational,
and engineering challenges required to make \AAP-based affordance
matching deployable in practice.
How the framework generalises to link-traversal, dereferencing, and hypermedia-driven access is left to future work.

The remainder of the paper is structured as follows.
Section~\ref{sec:background} revisits the structural insights of
\owls and \wsmo, and identifies which of them transfer to the \kg setting and which require formal extension.
This motivates Section~\ref{sec:dimensions}, which introduces the
four \AAP dimensions, including how each is evaluated against a task
signature and how the dimensions interact.
Section~\ref{sec:vision} grounds the framework in a worked example to illustrate how three superficially comparable \kgs result in distinct planning assessments.
Section~\ref{sec:research} then sets out a five-point research agenda required to make \AAP-based affordance matching deployable in practice before concluding in Section~\ref{sec:conclusions}.

\section{Agents, Knowledge and the SWS tradition}
\label{sec:background}

Two of the most influential formalisms to emerge from the work on
Semantic Web Services, \owls~\cite{ankolenkar2002damls,owls2004,martin2004owls}
and \wsmo~\cite{cabral2004approaches,WSMO2006}, embody a number of structural
insights regarding the formal description of resource and service capability
for agents~\cite{payne2008webservices}.  These insights equally hold for \kgs, and formed the basis for the motivation for an \emph{Agentic Affordance Profile}.
\owls organised service descriptions around three components addressing
three distinct questions that an agent must answer at different stages of
interaction. The \emph{Profile} answered \emph{``What does this service
do?''}, enabling \emph{discovery} by formal reasoning over the shared
ontologies of the requester and service
provider~\cite{paolucci2002matching}. The \emph{Process Model} answered
\emph{``How does it work internally?''}, enabling workflow composition and orchestration.
The \emph{Grounding} answered \emph{``How do I invoke it?''}, mapping
formal descriptions to protocol bindings.
The value of such an ontological grounding was demonstrated in domains where ontology-based representations enabled varying degrees of semantic interoperability, from lightweight ontology-assisted integration in early semantic grid systems~\cite{Wroe_Bioinformatics2004}, to explicit OWL-based ontology-mediated data access~\cite{hoehndorf2015aberowl}, and more recent ontology-driven harmonisation frameworks in materials science~\cite{ashino2010materialsontology,Bayerlein2024,morlidge2026}.
These same three questions arise for \kgs:
\begin{enumerate}[leftmargin=1cm,labelindent=0.5cm,topsep=2pt]
    \item 
        \emph{``What can I do with this \kg?''} -- what queries will return meaningful results, what reasoning is supported, and what task-relevant concepts are present;
    \item
        \emph{``How does querying work?''} -- what are the inference rules, update protocols, access constraints; and 
    \item
    \emph{``How do I query it?''} -- SPARQL endpoints, authentication, federation.
\end{enumerate}
Whilst current \kg metadata frameworks adequately answer the third question, they only answer the first at the content level, falling short
of the \owls \emph{Profile}-level question: \emph{what can this \kg do for an
agent with a specific task?} Answering this requires knowledge about what
reasoning is supported, whether the agent's task vocabulary is grounded
in the \kg's \textsc{TBox}, and what inferences can be formally trusted.
Therefore mirroring the \iope structure of \owls service profiles, a \kg's agentic affordance profile needs to describe three things from the agent's epistemic perspective:
1) the epistemic preconditions the agent must satisfy before engaging the \kg (e.g., task vocabulary grounded in the \kg's schema for meaningful queries);
2) what it can reliably conclude from the results of the query, and under what closure assumptions those conclusions hold; and 
3) how its knowledge base is extended by the interaction, expressed using the \kg's vocabulary.\footnote{
The scope of that extension is itself a design question: whether the agent
retains only assertions over its own signature, all consequences it can
derive, or the mediating vocabulary itself. We defer this to future work,
under the agent-to-agent generalisation identified in Section~\ref{sec:research}.}
Specifying these against shared domain ontologies enables formal
affordance matching, replacing the exploratory querying that agents
must currently perform: an approach that is costly, unreliable, and
epistemically risky.

Whereas \owls primarily addressed capability descriptions, \wsmo,
\emph{inter alia}, contributed a complementary insight: a typed taxonomy
of \emph{mediators}, treated as semantically specified bridges between
ontologically mismatched components~\cite{WSMO2006}.
These mediators were first-class objects: discoverable, composable, and
describable using the same vocabulary as services.
For \kg interoperability, the \wsmo mediator concept maps onto a formally specified \kg-to-\kg bridge: not just a functional adapter, but a transformation whose correctness can, in principle, be specified against a preservation criterion, enabling an agent to assess whether the transformed content remains adequate for its task.
However, registering a mediator simply as a static bridge is
insufficient for an agent operating under task-relative ontological
constraints, since it needs to determine if the mediator will close the
specific semantic gap for the task at hand \emph{before} it is invoked.
Mediators therefore need to declare formal input and output signatures,
grounded in the shared ontologies, so that their suitability can be assessed
at planning time rather than discovered through invocation.
Mediation must therefore be a planning object, not merely a resolution mechanism.

Existing linked-data quality and fitness-for-use
frameworks~\cite{debattista2018lodcloud,tartir2005ontoqa,zaveri2016survey},
including the \textsc{W3C} Data Quality Vocabulary
(\textsc{DQV})~\cite{debattista2016dqv},
characterise completeness, consistency and interpretability as
intrinsic properties of the \kg.
Competency questions (CQs)~\cite{CQs-Alharbi24} serve a related
purpose from a different perspective: they specify at design time the
questions that a computational resource is intended to support,
independent of any particular requester.
Gibson's ecological psychology~\cite{gibson1979ecological} defines an
\emph{affordance} not as a property of an environment alone but of
its relationship with an agent: what the environment \emph{offers}
relative to the agent's capacities and goals.
Applied to knowledge resources, a \kg's affordance is what it enables a \emph{specific agent}
to do on a \emph{specific task}, as opposed to what it
contains in the abstract.
Celino~\cite{celino2026ka} formalised this intuition as a
\emph{Knowledge Affordance (KA)} for knowledge resources, defined
relative to a requester's context, goals, and CQs.
This separates an affordance from a capability advertisement: a VoID
description describes the resource from the publisher's perspective
whereas an affordance describes it from the agent's; e.g., the same
\kg may have high affordance for a SNOMED-CT-grounded medical agent
and zero for a legal agent with no published alignment.
The \AAP differs from these approaches in two respects
(Table~\ref{tab:comparison}): it is \emph{relational}, scored against
a task signature computed at agent planning time rather than being
fixed at resource-design time or intrinsic to the resource alone; and
it is \emph{planning-actionable}, in that each dimensional shortfall
directly identifies a specific remedial action rather than merely
producing a quality assessment that leaves remediation unspecified.

\begin{table}[t]
    \centering
    \scriptsize
    \setlength{\tabcolsep}{5pt}
\caption{Existing frameworks compared with the proposed \AAP
along three axes: whether assessments are intrinsic to the \kg or
relational to a specific agent/task; whether task specificity is a
first-class input; and whether shortfalls map to specific remedial
actions at agent planning time.}
\label{tab:comparison}
\begin{tabular}{@{}llll@{}}
\toprule
\textbf{Framework} & \textbf{Intrinsic / Relational} & \textbf{Task-aware} & \textbf{Planning-actionable} \\
\midrule
\textsc{VoID}~\cite{alexander2011void}                & Intrinsic          & No               & No       \\
\textsc{DCAT}~\cite{maali2014dcat}                    & Intrinsic          & No               & No       \\
\textsc{SPARQL-SD}~\cite{williams2013sparql-sd}       & Intrinsic (endpoint) & No             & Partial  \\
\textsc{DQV}~\cite{debattista2016dqv}                 & Intrinsic          & No               & No       \\
OntoQA~\cite{tartir2005ontoqa}                        & Intrinsic          & No               & No       \\
Competency questions~\cite{CQs-Alharbi24}             & Design-time        & Yes              & No       \\
Knowledge Affordances~\cite{celino2026ka}              & Relational         & Yes (via CQs)    & Partial  \\
\midrule
\textbf{\AAP}                     & Relational & Yes ($\mathcal{S}$) & Yes (per-dimension) \\
\bottomrule
\end{tabular}
\end{table}

\looseness=-1
To frame this from the perspective of autonomous agents on
the Web, it is useful to distinguish \emph{signifiers} from
\emph{affordances}. Following the formalisation of
Vachtsevanou~et~al.~\cite{signifiers-Vachtsevanou23}, a \emph{signifier} is
a declared property of some resource, published for the agent to inspect
at discovery time, whereas an \emph{affordance} is a use that the agent can compute for
that specific resource relative to its own goals and capacities. \textsc{VoID},
\textsc{DCAT} and \textsc{SPARQL-SD} publish signifiers: metadata
declarations of dataset structure, cataloguing information, and
endpoint capabilities respectively. However, none of them publish
an affordance in the Gibsonian sense, as an affordance is not a
property of the resource alone, but a relation between resource and
agent, one the \AAP quantifies over the resource's signifiers
relative to the agent's task signature and epistemic requirements.
The \AAP is designed to fill this gap, such that for a specific class of agent and task, its four
dimensions score the resource's affordance over the publisher's
signifiers, so the agent-relative half of the
signifier/affordance decoupling is itself published in the metadata layer.
By publishing signifiers alone, current \kg metadata standards place the burden of affordance
computation entirely on the agent, at the cost of exploratory
querying whenever the signifiers under-determine fitness. An
\AAP-enabled registry inverts this: the affordance is
pre-computed against a shared task ontology and published alongside
the signifiers, so an agent's planning-time selection collapses to a
metadata query rather than exploratory interaction. Failure to make this affordance
computation available exposes the agent to \emph{epistemic failure}:
its plan is locally coherent given the signifiers it can inspect, yet
globally unsound as those signifiers do not warrant the task's
inferential or closure requirements.

\section{Agentic Affordances: A Four-Dimension Framework}
\label{sec:dimensions}
The \emph{agentic affordance} of a knowledge graph is characterised
across four dimensions: $\DimE$, $\DimG$, $\DimR$, and $\DimD$ (see Table \ref{tab:continuum}).
Each of these is a property of the \kg relative to a
specific agent and task, rather than of the \kg in isolation.
This distinguishes the \AAP from a capability advertisement or signifier, i.e.~not
what the resource is but what it offers a specific
requester, instantiating, at the agent level, the community-relative affordance framing of the ontological continuum~\cite{daga2026continuum}.
The continuum organises the \kg characterisation space around two
orthogonal distinctions: \emph{semantics} versus \emph{pragmatics}, and \emph{properties}
versus \emph{affordances}.
In this context, the \emph{semantic} axis concerns the meanings a
\kg encodes, while the \emph{pragmatic} axis concerns the mechanisms
by which those meanings are represented, encoded, and operated on.
Orthogonally, 
the \emph{properties} axis captures features that are intrinsic to the \kg
as an artefact, whereas the \emph{affordances} axis captures what the
\kg offers relative to a specific agent pursuing a specific task.

\begin{table}[ht]
\centering
    \setlength{\tabcolsep}{2pt}
    \caption{The \AAP dimensions positioned relative to the
Properties/Affordances $\times$ Semantic/Pragmatic structure of the
ontological continuum~\cite{daga2026continuum}.$^{\dag}$}
\label{tab:continuum}
\begin{tabular}{@{}lclcl@{}}
\toprule
 & \multicolumn{2}{l}{\textbf{Properties}} & \multicolumn{2}{l}{\textbf{Affordances}} \\
 & \multicolumn{2}{l}{\footnotesize\textit{(intrinsic to the \kg)}} 
 & \multicolumn{2}{l}{\footnotesize\textit{(relative to agent and task)}} \\
\midrule
\textbf{Semantic}  & & &   $\DimG$: & Task-Relative Grounding  \\
\footnotesize\textit{(meanings encoded)}
                   & &   & & (via signature closure) \\
\addlinespace
\textbf{Pragmatic} & $\DimE$: & Semantic Expressivity    & $\DimR$: & Epistemic Trust Scope   \\
\footnotesize\textit{(representational)} 
                   & &   (fragment\,+\,conformance)        & & (consistency\,+\,regime\,+\,scope) \\
\bottomrule
\end{tabular}
\par\smallskip
\begin{minipage}{0.95\linewidth}
{\footnotesize $^{\dag}$\,$\DimD$ (Agentic Discoverability) is a
meta-dimension: it measures what fraction of the affordance profile
above is pre-assessable from $M(\mathit{KG})$ alone, and therefore
sits outside this classification.}
\end{minipage}
\end{table}

The \AAP spans and operationalises the core of this space: $\DimG$ and $\DimR$ address, respectively, the semantic and pragmatic \emph{affordances}
that an agent requires from a \kg,
while $\DimE$ characterises the pragmatic properties of the schema.
$\DimD$ is \AAP-specific: it captures the epistemic cost of
the pre-interaction assessment. It has no direct counterpart to this dimension in the community-level continuum as it is relative to the
individual agent's discovery task.
Our main contribution is the argument that these dimensions are the principal axes along which agentic \kg affordances must be characterised rather than advocating any specific formalisation.  The formalism developed below is illustrative of one possible approach to operationalising them. A summary of notation used to define these dimensions appears later in Table~\ref{tab:notation}.

Each dimension can be viewed from two complementary perspectives (Table~\ref{tab:responsibilities}). From the publisher perspective, it determines what the \kg developer must expose, declare, or attest to. 
From the consumer perspective, it determines what must be evaluated against the agent's task before interaction. 
Furthermore, the role of the consumer is dependent on the type of interaction pattern assumed: for example, in a direct, peer-based setting, the consumer would be the agent itself; 
under Semantic-Web-Services-style matchmaking, the consumer may be shared by a matchmaking or discovery intermediary and the agent~\cite{keller2005wsmoDiscovery,paolucci2002matching};
and in link-traversal and hypermedia scenarios it would be distributed across successive resource evaluations~\cite{Ciortea2019Decade}.
However, the dimensions are agnostic over this choice of interaction pattern, as they only characterise the affordance with respect to some task.
From a multi-agent systems perspective, this division between publisher and consumer extends the signifier-vs-affordance distinction introduced in Section~\ref{sec:background}, whereby a \kg artefact exposes its signifiers, while an agent wanting to consume knowledge from that artefact for a given task expresses the relevant affordance~\cite{omicini2008aandartifacts}.
Increasing the metadata that the artefact precomputes and publishes
raises $\DimD$, reducing the agent's cognitive load when assessing
the viability of transacting with it, and the need to probe the artefact at runtime.
This publisher/consumer split is summarised across all four dimensions in Table~\ref{tab:responsibilities}.

\begin{table}[t]
    \centering
    \scriptsize
    \setlength{\tabcolsep}{4pt}
    \caption{Publisher/consumer split across the four AAP dimensions.
    The consumer column names what must be evaluated against the task; how that evaluation is distributed across matchmakers or the agent itself depends on the interaction pattern used.}
    \label{tab:responsibilities}
\begin{tabular}{p{2.2cm}p{4.4cm}p{4.5cm}}
\toprule
\textbf{Dimension} & \textbf{Publisher provides} &
                     \textbf{Consumer evaluates} \\
\midrule
$\DimE = (\mathcal{L}, \eta)$ &
  Fragment declaration for $\mathcal{L}$ (e.g., \textsc{OWL\,2} profile of $\mathcal{T}_{\mathit{KG}}$);
  conformance attestation for $\eta$. &
  Whether $\mathcal{L}$ supports the reasoning the task requires;
  whether $\eta$ meets the trust threshold. \\
\addlinespace
$\DimG$ &
  $\mathcal{T}_{\mathit{KG}}$; optionally a pre-computed
  $\mathcal{V}^{+}$ or definition-pattern catalogue. &
  Signature closure over task signature $\mathcal{S}$: is every
  $C \in \mathcal{S}$ in $\mathcal{V}^{+}$? \\
\addlinespace
$\DimR = (\rho_c,\rho_e,\rho_s)$ &
  Consistency certificate; entailment-regime declaration;
  predicate-level completeness statements. &
  Componentwise $\DimR \succeq \DimR_t^{\min}$ against the task's
  minimum requirements. \\
\addlinespace
$\DimD$ &
  Contents and completeness of $M(\mathit{KG})$
  (\textsc{VoID}, \textsc{DCAT}, \textsc{SPARQL-SD}, and any
  declared \AAP profile). &
  Which task types in $\mathcal{Q}$ can be pre-assessed from
  $M(\mathit{KG})$ without querying. \\
\bottomrule
\end{tabular}
\end{table}

\subsection{Semantic Expressivity ($\DimE$)} \label{sec:DimE}
The Semantic Expressivity ($\DimE$) dimension
characterises the expressivity of the ontology language governing the
\kg's schema, together with the degree to which the \kg content conforms
to that schema.
Without knowing the governing Description Logic (\textsc{DL}), an
agent cannot determine what it can soundly infer: for example, transitivity
reasoning requires support for transitive properties, while sound
query rewriting requires fragments such as \textsc{OWL\,2\,QL}.

\looseness=-1
This dimension is represented by the tuple $\DimE = (\mathcal{L},\eta)$.
The ontology language fragment $\mathcal{L}$ of the governing
schema is identified from the space of \textsc{RDF}, \textsc{RDFS},
the \textsc{OWL}~2 profiles (\textsc{EL}, \textsc{QL}, \textsc{RL}),
\textsc{OWL}~2 DL, and \textsc{OWL}~2 Full~\cite{baader2003dlhandbook,owl2profiles2012}.
These fragments do not form a single linear order: the \textsc{OWL}~2 profiles make different trade-offs (\textsc{EL} for polynomial-time classification, \textsc{QL} for query rewriting, \textsc{RL} for rule-based reasoning) and are not generally comparable by inclusion.
The task-side requirement $\mathcal{L}_t^{\min}$ is therefore expressed as a set of required constructors and inference capabilities; $\mathcal{L}(\mathit{KG}) \succeq \mathcal{L}_t^{\min}$ then holds when every constructor and inference capability required by task $t$ is supported by $\mathcal{L}(\mathit{KG})$.
This is paired with the \textsc{ABox} (instance) \emph{conformance} $\eta \in [0,1]$, defined as the fraction of \kg content well-typed under $\mathcal{T}_{\mathit{KG}}$, where well-typed means all class and property names used are declared in the governing \textsc{TBox} and their use is consistent with its typing constraints.
The conformance component $\eta$ complements the consistency-status
reading in $\DimR$ (Section~\ref{sec:DimR}). $\DimE$
characterises what the schema expresses, whereas $\DimR$ characterises
what the deployed endpoint actually returns.

Publishing $\DimE$ requires the \kg developer to declare $\mathcal{L}$
(for example, the \textsc{OWL\,2} profile of $\mathcal{T}_{\mathit{KG}}$, a
syntactic condition checkable with standard
tooling~\cite{owl2profiles2012}) and to determine $\eta$;
\textsc{SPARQL-SD} instead declares the endpoint's regime, which
belongs to $\rho_e$ in $\DimR$.
When consuming $\DimE$, the agent needs to verify that $\mathcal{L}$ supports the reasoning its task requires, and that $\eta$ meets its trust threshold.

\subsection{Task-Relative Grounding ($\DimG$)} \label{sec:DimG}
The Task-Relative Grounding ($\DimG$) measures the degree to which a \kg's vocabulary covers the agent's
\emph{task signature} $\mathcal{S}$ (i.e.\ the set of concept and role names required by the task).
Coverage not only includes explicitly declared entities in the \kg's vocabulary
but also implicit definability via the \kg's ontology axioms.
As different ontologies model the same domain at different
granularities, the required concepts may not appear directly in a
third-party \kg but may be expressible as complex class expressions, or be
derivable through its axiom structure.
In a knowledge-based agent framework, coverage is therefore a question
of formal inference over the \kg's theory, not lexical name matching: two \kgs may both expose an IRI named
$\mathsf{Invited\_speaker}$ and return populated query results, yet
those results carry different epistemic weight. 

For example, in \textsc{Conference.owl},\footnote{Part of the OAEI conference
dataset (\url{https://oaei.ontologymatching.org}). The worked example
(Section~\ref{sec:vision}) uses \textsc{Conference}$^{+}$: an extension
of the \textsc{Conference.owl} ontology that adds the concept and
role names $\mathsf{Researcher}$, $\mathsf{authorOf}$, and
$\mathsf{givenAt}$.} $\mathsf{Invited\_speaker}$
is defined by an explicit equivalence axiom ($\mathsf{Invited\_speaker}
\equiv \exists\mathsf{contributes}.\mathsf{Invited\_talk}$), and hence is
Beth-definable from a sub-signature of the TBox's
vocabulary~\cite{geleta2016ekaw}.
In a second \kg using the same IRI only in its \textsc{ABox}, the
name $\mathsf{Invited\_speaker}$ is present but its meaning is undefined by the schema; the
results match the label, but 
there is no commitment about what the label denotes in
the \kg, and hence it carries no semantic guarantee.
Dimension~$\DimG$ captures this distinction: a \kg grounds concept
$C$ if and only if $C$ is explicitly axiomatised or implicitly
definable from a \textsc{TBox} sub-signature.

A concept $C \in \mathcal{S}$ is \emph{implicitly definable} from
a signature $\Sigma \subseteq \mathrm{Sig}(\mathcal{T})$ under a
TBox $\mathcal{T}$ if any two models of $\mathcal{T}$ that agree
on $\Sigma$ also agree on $C$ (Beth definability~\cite{bethdef}).
Intuitively, $C$ is implicitly definable from $\Sigma$ if the axioms in
$\mathcal{T}$ leave $C$ no freedom once the meanings of the $\Sigma$-symbols are fixed.

Let $\mathcal{V} \subseteq \mathit{Sig}(\mathcal{T}_{\mathit{KG}})$ denote the \emph{exposed signature} of the \kg (the set of concept and role names the publisher makes directly accessible to consumers).
In the simplest case, the whole \textsc{TBox} signature is exposed and $\mathcal{V} = \mathit{Sig}(\mathcal{T}_{\mathit{KG}})$; where the publisher exposes only a modular subset, $\mathcal{V}$ is a proper subset.
The \emph{coverage problem} asks whether every $C \in \mathcal{S}$ is either in $\mathcal{V}$ or implicitly definable from $\mathcal{V}$ via $\mathcal{T}_{\mathit{KG}}$ in the Beth sense~\cite{geleta2016ekaw,geleta2016owled}.
We therefore define:
\begin{equation}
  \DimG = \frac{|\{C \in \mathcal{S} : C \in \mathcal{V}^{+}\}|}{|\mathcal{S}|}
\end{equation}
where $\mathcal{V}^{+}$ is the \emph{signature
closure}~\cite{geleta2016owled}: the set of concept and role names
either in $\mathcal{V}$ or implicitly definable from $\mathcal{V}$
via $\mathcal{T}_{\mathit{KG}}$ in the Beth sense. When
$\mathcal{V} = \mathit{Sig}(\mathcal{T}_{\mathit{KG}})$, the closure
reduces to $\mathcal{V}$ itself, whereas when $\mathcal{V}$ is a proper
subset, Beth-definability provides the mechanism for extending it
with concepts derivable from $\mathcal{V}$ via
$\mathcal{T}_{\mathit{KG}}$'s axioms.

While Beth definability provides the \emph{semantic} criterion for membership in $\mathcal{V}^{+}$, this membership can be decided by a standard subsumption check under a renaming of the non-$\mathcal{V}$ symbols, in the style of Padoa's method~\cite{tencate2013beth}.
$\mathcal{V}^{+}$-membership is therefore decidable in every fragment considered here, at the complexity of subsumption in that fragment.
Producing an explicit $\mathcal{V}$-definition of $C$, which the agent needs in order to query for $C$ directly, is a separate matter. 
For \textsc{OWL\,2\,QL}/\textsc{DL-Lite}, perfect query rewriting~\cite{calvanese2007obda} can be used. In \textsc{OWL\,2\,EL}, module extraction can be used as a sound pre-filter~\cite{doran2009aaai,cuencagrau2008modular}. 
For more expressive fragments, uniform interpolation provides the general machinery~\cite{lutz2011interpolation}, with the caveats that uniform interpolants need not exist~\cite{lutz2011interpolation} and that explicit definitions are guaranteed only in fragments with the Beth definability property~\cite{tencate2013beth}.
The \textsc{RDFS} case is simpler, though not by reachability alone: \textsf{rdfs:subClassOf} and \textsf{rdfs:subPropertyOf} axioms are one-directional, so a subsumption chain from a name to $\mathcal{V}$ bounds its interpretation without fixing it. A name outside $\mathcal{V}$ is definable from $\mathcal{V}$ exactly when it is provably mutually subsumed with a $\mathcal{V}$-name, so membership in $\mathcal{V}^{+}$ is computable in polynomial time from the strongly connected components of the declared subsumption graph, with no theorem-proving overhead. \textsf{rdfs:domain} and \textsf{rdfs:range} axioms constrain instance typing but, being one-directional, do not extend $\mathcal{V}^{+}$.
KG$_1$ in the worked example (Section~\ref{sec:vision}) is a degenerate instance: its schema declares only $\mathsf{Researcher}$, $\mathsf{Paper}$, and $\mathsf{authorOf}$, with no subsumption axioms, so $\mathcal{V}^{+} = \mathcal{V}$ and the coverage check is direct membership.

Definition-pattern catalogues, however, can offer a complementary, largely DL-independent route that
covers many real-world cases without invoking these general
techniques~\cite{geleta2016ekaw}.
Irrespectively of how this is computed, $\mathcal{V}^{+}$ may either be
computed by the publisher and shipped as part of $M(\mathit{KG})$,
or by the consumer at planning time against
$\mathcal{T}_{\mathit{KG}}$. The choice trades publisher effort
against consumer assessment cost, and directly influences the
$\DimD$ value the same \kg presents to different consumers.

This dimension is task- and agent-relevant: the same \kg may fully
ground one task's signature whilst not covering that of another.
When $\DimG < 1$, the gap identifies exactly which concepts require
mediation or knowledge acquisition, connecting the framework directly
to on-demand alignment~\cite{geleta2016ekaw,agentom2025} or \wsmo-style mediation~\cite{WSMO2006}.
$\DimG$ is a schema-level metric: it assesses whether the \kg's
vocabulary can ground the task concepts, not whether relevant instances
exist; instance-level relevance is only partially addressed by
\textsc{VoID} entity statistics.

\subsection{Epistemic Trust Scope ($\DimR$)} \label{sec:DimR}
\looseness=-1
This is a structured profile characterising when an agent can treat
query results as actionable inferences, determined by the \kg's world-closure
assumptions, completeness declarations, and consistency status.
In \owls and \wsmo, service descriptions were grounded in explicit
formal semantics, enabling agents to reason about the advertised functional properties
of a service (i.e. the \emph{service capabilities}) prior to invocation
and to determine what they could soundly \emph{conclude}
after invocation; whereas current \kg metadata standards do not provide such a comprehensive
or operational account of entailment or closure.
The critical question for $\DimR$ is therefore not only what the \kg
can express (Dimension~$\DimE$) but what can be inferred from an
\emph{empty or partial} result, which depends on the \kg's
\emph{world-closure assumption} and any \emph{predicate-level
completeness declarations}.

Consider a medical agent querying for contraindications between two drugs.
Under the closed world assumption (CWA), as in a relational database
where the contraindications table is assumed complete,
an empty result is a sound negative conclusion: no contraindication exists.
Under the open world assumption (OWA), as in an \textsc{OWL} \kg with no
completeness assertion, an empty result means only that no
contraindication is \emph{recorded}; the agent cannot safely conclude
absence, and any plan predicated on that conclusion is epistemically
unsound.
The Locally Closed World Assumption (\textsc{LCWA})~\cite{etzioni1997lcwa} applies \textsc{CWA} selectively to specific predicates, declarable for \textsc{RDF} via completeness statements~\cite{darari2013completeness}. \textsc{SHACL}~\cite{knublauch2017shacl} supplies closed-world validation, but \textsf{sh:closed} constrains a node's properties rather than a predicate's extension, and so it cannot by itself warrant negative conclusions.
$\DimR$ must therefore characterise not only whether a \kg operates
globally under \textsc{OWA} or \textsc{CWA}, but which predicates or
shapes are locally closed.
Thus, $\DimR = (\rho_c, \rho_e, \rho_s)$ is a three-component profile characterising which
closure assumptions apply:
\begin{enumerate}[leftmargin=1cm,labelindent=0.5cm,topsep=0pt,label=(\alph*)]
    \item 
        \textbf{consistency status}\,$\rho_c$\,(uncertified\,\!/\,\!TBox-consistent\,\!/\,\!jointly consistent);\footnote{\emph{Uncertified}: no consistency check has
been performed. \emph{TBox-consistent}: the schema alone is verified
consistent. \emph{Jointly consistent}: both  TBox and ABox are
verified consistent~\cite{baader2003dlhandbook}.}
    \item 
        \textbf{declared entailment regime} $\rho_e$ (none\,/\,\textsc{RDFS} / \textsc{OWL\,2\,EL} / \textsc{OWL\,2} \textsc{QL} / \textsc{OWL\,2\,RL} / \textsc{OWL\,2\,DL}); 
    \item 
        \textbf{completeness scope} $\rho_s$: the set of task-relevant predicates for which closure is declared, with semantics of global \textsc{CWA}, predicate-level \textsc{LCWA}~\cite{etzioni1997lcwa,motik2010bridging}, or explicit completeness statements~\cite{darari2013completeness}.
\end{enumerate}
Given a task's minimum epistemic requirement $\DimR_t^{\min} = (\rho_{c,t}^{\min}, \rho_{e,t}^{\min}, \rho_{s,t}^{\min})$, the relation $\DimR(\mathit{KG}) \succeq \DimR_t^{\min}$ used in the feasibility predicate for task $t$ holds iff $\rho_c(\mathit{KG}) \succeq \rho_{c,t}^{\min}$ (natural ordering on consistency levels),
$\rho_e(\mathit{KG}) \succeq \rho_{e,t}^{\min}$ (the declared regime
supports the constructors and inferences the task requires, as defined
in Section~\ref{sec:DimE}), and $\rho_s(\mathit{KG}) \supseteq
\rho_{s,t}^{\min}$ (set containment on closure-scope predicates).
Thus, $\DimR$ is not scalar trust but a structured epistemic adequacy
profile: its three components measure qualitatively different
concerns and are ordered independently.
In practice, $\DimR$ should be derived from a combination of declared
metadata (\textsc{SPARQL-SD} entailment regime declarations,
\textsc{SHACL} shape graphs), endpoint configuration, and, for
consistency status, external certification or runtime verification.

$\DimR$ determines what an agent can \emph{infer} from a \kg, not merely
what it can retrieve.
For example, safety-critical or regulatory reasoning may require an entailment regime
of at least \textsc{OWL\,2\,DL} and completeness declarations for the
relevant predicates~\cite{motik2010bridging}; scientific classification may require only \textsc{OWL\,2\,EL}~\cite{hoehndorf2015aberowl,owl2profiles2012} with consistency certification; whereas factual retrieval
operates soundly under \textsc{SPARQL} Simple entailment~\cite{glimm2013sparqler}, provided the
agent draws no negative conclusions.
However, if $\DimR$ is undeclared, an agent that conflates the
open and closed world assumptions may produce plans that are locally
coherent but globally unsound: an epistemic failure invisible to the
other three dimensions.
A \kg may have an \textsc{OWL\,2\,DL} governing schema but be deployed
as a plain triple store, leaving its operative entailment regime at \textsc{SPARQL} Simple entailment only.
Conversely, a plain \textsc{RDFS} \kg whose publishers declare
completeness over a restricted predicate set provides a well-defined,
non-trivial affordance profile for queries over that set.
However, many real-world \kgs make heavy use of undefined
vocabulary~\cite{debattista2018lodcloud}, and often operate without
engaging their reasoning capacity or making the completeness
declarations that would make query results epistemically actionable.

The three components of $\DimR$ are handled separately on both sides.
A publisher populates each through a different mechanism: $\rho_c$
via consistency certification, $\rho_e$ via entailment-regime
declaration, and $\rho_s$ via completeness statements. A consumer
checks each independently against the task's minimum, which is why
$\succeq$ is defined for each component separately.

\begin{sloppypar}
The current definition assumes a single per-\kg $\DimR$ tuple; however, real \kgs may
exhibit heterogeneous consistency, entailment regimes, or completeness
scopes across named graphs~\cite{carroll2005namedgraphs} or federated
sub-endpoints. Generalising $\DimR$ from a single tuple to a set of
context-scoped tuples is a natural extension that is discussed in more detail in Section~\ref{sec:research}.
\end{sloppypar}

\subsection{Agentic Discoverability ($\DimD$)} \label{sec:DimD}
Agentic Discoverability ($\DimD$) is qualitatively different from
$\DimE$, $\DimG$, and $\DimR$. Where they characterise
substantive affordance conditions (i.e.\ what the \kg supports),
$\DimD$ characterises the \emph{extent} to which those affordances can be
established from the declarations available prior to interaction.
It is therefore not a property the \kg affords, but a property of
the affordance-assessment process.
Current \kgs already occupy positions along this scale:
\textsc{VoID}, \textsc{DCAT}, and \textsc{SPARQL-SD} declarations
partially reduce assessment cost by exposing properties
relevant to $\DimE$ and $\DimR$ at the metadata level, while
a full \AAP profile would occupy the upper bound of this scale.
A \kg at the lower end would require exploratory queries before the agent
can assess its fitness for the desired task, whereas one at the upper
end can be evaluated at planning or composition time from its metadata
alone, enabling rational \kg selection without committing to engage a
resource.
The characteristics for this dimension span from no description (raw
\textsc{SPARQL} endpoint) $\to$ structural and capability metadata
(\textsc{VoID}~\cite{alexander2011void},
\textsc{DCAT}~\cite{maali2014dcat},
\textsc{SPARQL-SD})~\cite{williams2013sparql-sd}
$\to$ natural language documentation $\to$ a full \AAP profile, providing formal affordance scores over supported task types, that are grounded
in shared domain ontologies.

A \kg is \emph{fit} for some task $t$ when it satisfies all four
feasibility requirements: the \emph{fragment} ($\mathcal{L}$), \emph{conformance} ($\eta$), \emph{grounding} ($\DimG$), and
\emph{regime} ($\DimR$) conditions formalised in the feasibility predicate of
Section~\ref{sec:discussion}.
$\DimD$ measures the fraction of this feasibility condition that is
decidable from $M(\mathit{KG})$ alone (i.e., from metadata, without
querying the \kg). Formally, let $M(\mathit{KG})$ denote the set of
metadata assertions about $\mathit{KG}$, treated as a formal
\textsc{RDF}/\textsc{OWL} theory (\textsc{SPARQL-SD} endpoint
declarations, \textsc{VoID}/\textsc{DCAT} descriptors, and any
declared \AAP profile), and let $\mathit{fitness}(\mathit{KG},t)$
denote the conjunction of these four requirements for task type $t$.
We then define:
\begin{equation}
  \DimD(\mathit{KG}) \;=\; \frac{|\{\,t \in \mathcal{Q} \;\mid\;
    M(\mathit{KG}) \text{ decides } \mathit{fitness}(\mathit{KG},\,t)\,\}|}{|\mathcal{Q}|}
\end{equation}
where $\mathcal{Q}$ is the set of agent task types defined at design
time against a shared task ontology (itself a prerequisite identified
in Section~\ref{sec:research}).\footnote{
Note that $\mathcal{Q}$ is a set of task \emph{classes}, not individual queries,
so the AAP's scaling concern reduces to the task-ontology-sizing question
identified above rather than to combinatorial enumeration of possible queries.}
$M(\mathit{KG})$ \emph{decides} $\mathit{fitness}(\mathit{KG},t)$ when it
entails either $\mathit{fitness}(\mathit{KG},t)$ or its negation.
Metadata that rules a \kg out for a task spares the agent the need for
exploratory querying, just as it does when it rules the \kg in;
$\DimD$ thus measures the availability of a verdict, not its polarity.
A precomputed $\DimD$ characterises whether the information required to assess fitness for a task class in $\mathcal{Q}$ is available from $M(\mathit{KG})$ alone. 
Evaluation of a particular task instance, whose signature may differ in detail from the class template, is performed at planning time against that instance's specific signature (Section~\ref{sec:vision}).
A score of 1 means the \kg's metadata suffices to pre-assess the
suitability for every task type; a score of 0 means the metadata decides none of them.
This assumes that the metadata is accurate and current; in noisy or
adversarial settings, $\DimD$ should be interpreted relative to
trusted or independently certified metadata sources.

\begin{table}[t]
    \centering
    \scriptsize
    \setlength{\tabcolsep}{5pt}
    \caption{Notation used throughout the paper, grouped by dimension.}
    \label{tab:notation}
    \begin{tabular}{l@{\hspace{1em}}l@{\hspace{0.5em}}p{9.6cm}}
\toprule
\multicolumn{3}{@{}l@{}}{\textbf{Semantic Expressivity}: \textit{tuple $(\mathcal{L}, \eta)$ pairing DL fragment with \textsc{ABox} conformance}} \\
  \addlinespace
\multirow{4}{*}{$\DimE$~~~~}
  & $\mathcal{L}$ & DL fragment of the governing ontology \\
  & $\eta$ & \textsc{ABox} conformance: fraction of \kg content well-typed under $\mathcal{L}$ \\
  & $\mathcal{L}_t^{\min}$ & Minimum DL fragment required by task $t$ \\
  & $\eta_t^{\min}$ & Minimum \textsc{ABox} conformance required by task $t$ \\
\midrule
\multicolumn{3}{@{}l@{}}{\textbf{Task-Relative Grounding}: \textit{vocabulary coverage of $\mathcal{S}$ via $\mathcal{V}^{+}$}} \\
  \addlinespace
\multirow{4}{*}{$\DimG$}
  & $\mathcal{S}$ & Task signature: set of concept and role names required by a task \\
& \multirow{2}{*}{$\mathcal{V}$}        & Exposed signature $\subseteq \mathit{Sig}(\mathcal{T}_{\mathit{KG}})$: names the publisher makes accessible (typically the full \textsc{TBox} signature) \\
 & $\mathcal{V}^{+}$ & Signature closure: names in $\mathcal{V}$ or Beth-definable from $\mathcal{V}$ under $\mathcal{T}_{\mathit{KG}}$ \\
 & $\mathcal{T}_{\mathit{KG}}$ & \kg's \textsc{TBox} (including any imported ontologies) \\

\midrule
\multicolumn{3}{@{}l@{}}{\textbf{Epistemic Trust Scope}: \textit{tuple $(\rho_c, \rho_e, \rho_s)$ pairing consistency, entailment, and closure}} \\
  \addlinespace
\multirow{7}{*}{$\DimR$}
  & $\rho_c$ & Consistency status (first component of $\DimR$) \\
  & $\rho_e$ & Declared entailment regime (second component of $\DimR$) \\
  & $\rho_s$ & Completeness scope (third component of $\DimR$) \\
  & $\rho_{c,t}^{\min}$ & Minimum consistency status required by task $t$ \\
  & $\rho_{e,t}^{\min}$ & Minimum entailment regime required by task $t$ \\
  & $\rho_{s,t}^{\min}$ & Minimum closure scope required by task $t$ \\
  & $\DimR_t^{\min}$ & Minimum epistemic requirement for task $t$: tuple $(\rho_{c,t}^{\min}, \rho_{e,t}^{\min}, \rho_{s,t}^{\min})$ \\
\midrule
\multicolumn{3}{@{}l@{}}{\textbf{Agentic Discoverability (meta-)dimension}: \textit{fraction of fitness pre-assessable from $M(\mathit{KG})$}} \\
  \addlinespace
\multirow{2}{*}{$\DimD$}
  & $\mathcal{Q}$        & Set of agent task classes defined against a shared task ontology \\
  & $M(\mathit{KG})$     & Metadata about the \kg (VoID / DCAT / SPARQL-SD / \AAP declarations) \\
\bottomrule
\end{tabular}
\end{table}

$\DimD$'s ceiling is set by publisher choices: the metadata
vocabularies used, and the completeness of the resulting
$M(\mathit{KG})$. Its value depends on the consumer's task
portfolio $\mathcal{Q}$, and is exploited differently across
interaction patterns; for example, an agent avoids probing or a discovery service
builds a richer index. The dimension itself measures
the \kg's contribution to any of these, independently of the choice of the
consumer architecture.

\begin{figure}[t]
    \centering
    \includegraphics[width=0.95\textwidth,
                 trim=10cm 9.2cm 11cm 7.5cm,
                 clip]{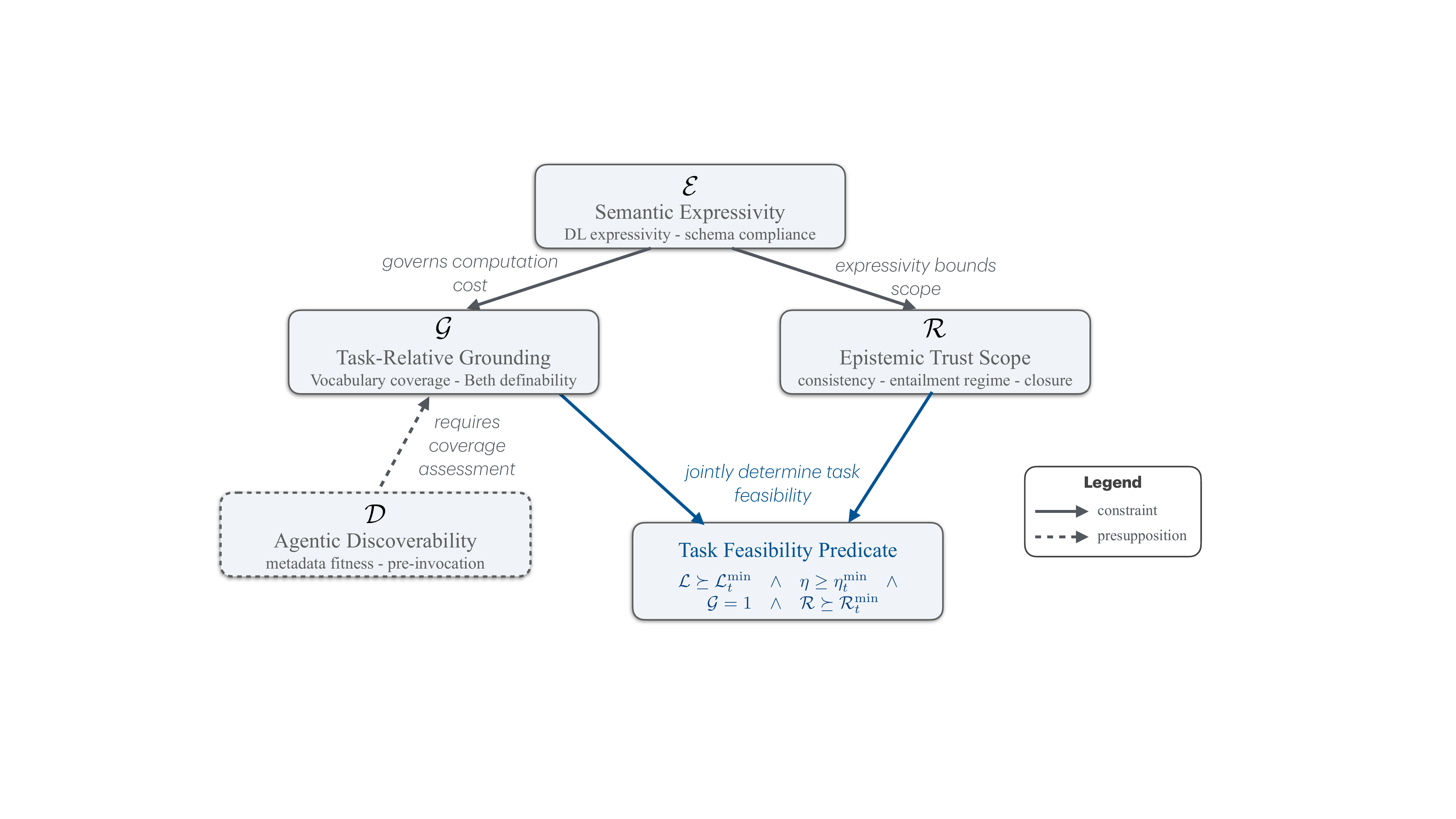}
\caption{Interaction structure of the four \AAP dimensions.
$\DimE$ constrains the entailment-regime component $\rho_e$ of
$\DimR$ (bounding the maximum expressible regime) and the
computability cost of $\DimG$ (the tractability of the coverage
check depends on the governing \textsc{DL} fragment).
$\DimD$ is a meta-dimension: it presupposes $\DimG$ and is bounded
by the fraction of $\DimR$-relevant facts that $M(\mathit{KG})$
actually publishes.
$\DimG$ and $\DimR$ jointly determine task feasibility, in two
layers: coverage of the task signature, and sufficiency of the
closure assumptions for sound negative inference.}
\label{fig:dimensions}
\end{figure}

\subsection{Discussion}\label{sec:discussion}

\medskip
\noindent\textbf{Interactions:~}
The dimensions form a partially ordered
affordance space, with three interactions that are particularly
important, as illustrated in Figure \ref{fig:dimensions}.
\begin{enumerate}[leftmargin=1cm,labelindent=0.5cm,topsep=0pt]
    \item 
$\DimE$ \emph{constrains} the entailment-regime component $\rho_e$ of $\DimR$ and the computability of $\DimG$.
For $\rho_e$, the constraint is on \emph{scope}: as the reasoning
regime cannot exceed the vocabulary expressivity, the maximum regime
that a \kg can usefully declare is bounded by $\DimE$. For $\DimG$, the constraint is on \emph{cost}.  
Module extraction and uniform interpolation are tractable only for specific \textsc{DL} fragments,
and thus the efficiency of this coverage check will vary with the governing schema. 
Whether $\DimG$ can be assessed at all is independent of $\DimE$:
the task signature $\mathcal{S}$ is simply a set of concept and role
names, with no DL fragment of its own. As such, $\DimG$ can be
assessed against any task regardless of the KG's expressivity level
(though the cost will be expressivity-dependent).
    \item
$\DimD$ depends on $\DimG$ \emph{and} $\DimR$. 
With respect to $\DimG$, the dependency is a \emph{precondition}. An
agent that cannot assess coverage of its task signature cannot use
even a richly described \kg, so $\DimG$ must hold for $\DimD$ to be
informative at all. With respect to $\DimR$, the dependency is on
\emph{publication}. Without closure declarations being part of what
$M(\mathit{KG})$ records, the agent cannot determine at planning time
whether negative conclusions will be sound. Thus, the value of $\DimD$ is bounded by
the fraction of $\DimR$-relevant facts the metadata actually
publishes.
    \item 
$\DimG$ and $\DimR$ \emph{jointly determine task feasibility}. 
An agent needs coverage of task concepts ($\DimG = 1$) \emph{and} an entailment regime at least equal to that required by the task.
For tasks that require \emph{sound negative conclusions}, the corresponding predicates must also be declared closed under $\DimR$, as a grounding alone is insufficient if the
\kg's closure assumptions do not permit absence reasoning.
\end{enumerate}
Thus, the four dimensions jointly determine whether a task $t$ is
(i)~\emph{representable}: whether the \kg's \textsc{DL} fragment
supports the required inference ($\DimE$);
(ii)~\emph{grounded}: whether the task vocabulary is covered by the
\kg's signature closure ($\DimG$);
(iii)~\emph{epistemically sound}: whether query results can be treated
as actionable inferences under appropriate closure assumptions ($\DimR$);
and (iv)~\emph{selectable without exploration}: whether these properties
are decidable from $M(\mathit{KG})$ alone, prior to interaction ($\DimD$).

\medskip
\noindent\textbf{Feasibility predicate.~}
The feasibility of a task given a \kg can be determined through the use of a single planning-actionable
predicate without the need for exploratory invocation. Given task $t$ with signature
$\mathcal{S}_t$, minimum fragment $\mathcal{L}_t^{\min}$, minimum
\textsc{ABox} conformance $\eta_t^{\min}$, and minimum epistemic
requirement $\DimR_t^{\min} = (\rho_{c,t}^{\min}, \rho_{e,t}^{\min}, \rho_{s,t}^{\min})$
(Section~\ref{sec:DimR}), $t$ is \emph{feasible} against $\mathit{KG}$ iff all
four dimensional feasibility requirements are satisfied:
\begin{equation}
    \label{eq:feasibility}
\begin{aligned}
\text{fragment requirement:} & \quad \mathcal{L}(\mathit{KG}) \succeq \mathcal{L}_t^{\min} \\
\text{conformance requirement:} & \quad \eta(\mathit{KG}) \geq \eta_t^{\min} \\
\text{grounding requirement:} & \quad \DimG(\mathit{KG}, \mathcal{S}_t) = 1 \\
\text{regime requirement:} & \quad \DimR(\mathit{KG}) \succeq \DimR_t^{\min}
\end{aligned}
\end{equation}
The DL fragment $\mathcal{L}$ additionally bounds the computability
of the $\DimG$ check and the maximum regime expressible in $\DimR$,
while $\DimD$ determines whether the agent can evaluate these requirements
\emph{at planning time} from $M(\mathit{KG})$ alone, or must resort
to exploratory querying. 
The strictness of the grounding requirement ($\DimG = 1$) reflects the
case in which no mediation is required. Agents that fail this requirement may subsequently and
successfully invoke a knowledge-producing service to bridge missing
concepts (Section~\ref{sec:vision}), thereby raising their operative
$\DimG$ before re-evaluation.

As each requirement corresponds directly to a specific dimension, a failing
requirement is diagnostic, since it identifies the dimension responsible. By using 
the \emph{dimension-to-action} mapping (Section~\ref{sec:vision}), an agent can consider the appropriate class of remedy (vocabulary mediation, \kg re-selection,
or schema/content repair) prior to any future invocation.
This factorisation is what gives the \emph{affordance gap signal} its
diagnostic power.

\medskip
\noindent\textbf{Novelty over \owls/\wsmo:~}
The \AAP makes three advances over \owls and \wsmo.
They differ in formal depth: the first introduces a criterion absent
from service-profile matching; the second and third reframe existing
insights for the deployed \kg setting.
Coverage in $\DimG$ is \emph{definability}-based, not alignment-based: it does not require explicit ontology correspondences between the agent's vocabulary and the \kg's schema. 
Instead, implicitly derivable concepts are grounded via signature closure and Beth definability, a criterion absent from \owls and \wsmo, both of which instead determined coverage through subsumption-based matching over shared ontologies~\cite{paolucci2002matching} or explicitly declared mediators~\cite{WSMO2006}.
Furthermore, $\DimE$ and $\DimR$ are \emph{independently variable}: unlike the formulation proposed here, \owls and \wsmo did not isolate the divergence between a resource's governing ontology language and its operative entailment regime as an explicit dimension of resource fitness.  However, in deployed \kgs these two routinely diverge. A \kg may declare \textsc{OWL\,2\,DL} and yet be queried as a plain triple store without inference, leaving an agent unable to detect that its declared reasoning capacity is not operationally available.
Finally, all scores are \emph{agent- and task-relative}: the same \kg may result in different scores for $\DimG$ across different tasks, instantiating the Gibsonian affordance framing at the formal level~\cite{celino2026ka,gibson1979ecological}. 
This decomposition is operationalised in a way that is absent from service profile matching: each dimensional shortfall identifies a class of candidate remedial actions (e.g., vocabulary mediation for a $\DimG$ failure, \kg re-selection for a $\DimR$ failure) rather than producing a generic quality judgement.

\section{Agentic Affordance Profile in Practice}
\label{sec:vision}

We envisage utilising this four-dimension framework 
via an \emph{Agentic Affordance Profile} (\AAP) for \kgs,
functioning as a semantic layer above VoID and DCAT. In the
signifier/affordance framing introduced in
Section~\ref{sec:background}, this places the \AAP as the
affordance layer, pre-computed and published alongside the signifier
metadata that those standards already provide. 

An \AAP is not a single universal score attached to a \kg but a
structured profile containing resource-level declarations for $\DimE$
and $\DimR$, together with pre-computed $\DimG$
and $\DimD$ assessments for a declared family of agent task classes.
A consumer may then apply these assessments against
its own specific task signature. The profile is expressed using a
vocabulary defined in \textsc{OWL} and grounded in shared domain
ontologies, and publishing it in a \kg registry, for example, would
enable formal affordance matching at agent planning time.
An \AAP is thus not a capability advertisement. Rather than
describing what the \kg contains from the publisher's
perspective, it describes what a specific class of agent can
do with the \kg. This includes what the agent can prove, what
it can safely conclude from an empty result, and where mediation is
required before interaction is epistemically sound.

\subsection{The Knowledge Cycle}\label{sec:cycle}

The \AAP assumes that agents behave \emph{rationally} with
respect to \kg discovery. Before committing to engage with some \kg, an agent
evaluates whether the \kg can further its goals in a pre-invocation phase
we term the \emph{knowledge cycle}.
The cycle has three phases: \emph{discovery}, \emph{plan formation}, and (when
multiple \kgs are involved) \emph{composition}. 
We adopt the Belief-Desire-Intention (\textsc{BDI}) model as the reference deliberation model~\cite{wooldridge1995intelligent}, as \textsc{BDI} makes the belief-update, goal-checking, and intention-commitment steps in which \AAP scores are consumed explicit. The framework is, however, architecture-agnostic: an LLM-orchestrated planner could also consume \AAP scores directly via a registry tool wrapper, treating them as preconditions on tool-use without a classical \textsc{BDI}
loop~\cite{malfa2025large}.
The description below assumes a registry-based matchmaking pattern;
the same phase structure applies to other consumer patterns
(peer-based direct assessment, hypermedia-driven navigation, etc.), with
the registry replaced by the relevant discovery mechanism.

At \emph{discovery} time, an agent queries an \AAP-enabled
registry for \kgs whose expressivity meets the task minimum and whose entailment regime is sufficient,
identifying candidates without requiring content access. This is analogous to \owls \emph{Profile}
matching~\cite{paolucci2002matching} applied to \kgs.

During \emph{plan formation}, the agent performs a coverage check
to determine whether every concept in its task signature $\mathcal{S}$
is included in the \kg's signature closure $\mathcal{V}^{+}$
($\DimG = 1$).
Unlike KG expressivity and the entailment regime, coverage depends on the specific
task signature and so is checked against candidate \kgs at planning
time rather than at discovery.
If $\DimG < 1$, the agent determines which concepts in $\mathcal{S}$ lie
outside $\mathcal{V}^{+}$ and searches for knowledge-producing services
(such as \psms~\cite{motta1999reusable}) that can supply the missing
knowledge.
These may include dynamic ontology alignment procedures, SPARQL
CONSTRUCT queries over external \kgs, or inference services grounded in
the \kg's axioms.
The results of invoking these services update the agent's epistemic state, thereby supplying the missing knowledge. The coverage check is then
repeated.

Once coverage is confirmed, the agent can proceed to contact the \kg.
\emph{Composition} introduces a further consideration. If an agent's
tasks involve querying multiple \kgs, or if a pre-registered
mediator is used to address a conceptual gap, an \AAP-based
composition check determines whether the composite task signature
$\mathcal{S}$ is covered by the union of signature closures
$\mathcal{V}^{+}$.\footnote{
Union coverage of $\mathcal{V}^{+}$ is a first approximation;
Section~\ref{sec:research} discusses why semantically coherent
composition requires additional coherence conditions on the
participating \kgs' entailment regimes and closure declarations.
}
Where gaps exist, the framework identifies which concepts require
alignment or mediation; both classical and \textsc{LLM}-driven approaches
to on-demand alignment are
applicable~\cite{geleta2016ekaw,jimenezruiz2016limiting,agentom2025,zhang2024aamas}.
The composition case poses a particular efficiency challenge:
naively, every correspondence in every aligned ontology is a
candidate for closing the gap. The use of module extraction over the task
signature~\cite{doran2009aaai} can provide a pre-invocation filter:
only correspondences whose source entities lie within the
task-scoped module will be candidates for gap resolution, and where
no such correspondences exist, alignment failure can be predicted
without invoking any negotiation service.
A registered mediator can thus be evaluated against the specific gap
$\mathcal{S} \setminus \mathcal{V}^{+}$ before any invocation is
committed. Thus, mediation becomes a \emph{planning object}:
something the agent selects and commits to at planning time, rather than
being invoked opportunistically at runtime.

When \emph{troubleshooting} task failure, current frameworks typically
offer no structured account of why a task failed: no result is received by the agent, or worse, an unexpected, inaccurate, or invalid result is received, with no information about which property
of the \kg was responsible for the failure.
However, this affordance framework maps a task failure into a principled diagnosis of the problem by attributing it to a specific dimensional gap:

\begin{enumerate}[leftmargin=1cm,labelindent=0.5cm,topsep=0pt]
    \item 
        \emph{where task concepts lie outside the \kg's signature closure} $\mathcal{V}^{+}$ (a $\DimG$-failure), the \kg lacks the vocabulary to ground the task.  Candidate remedies include vocabulary mediation or the invocation of a knowledge-producing service to supply the missing grounding;
    \item 
        \emph{if the \kg's entailment regime or closure assumptions fall short of what the task requires} (a $\DimR$-failure), the issue lies not in the \kg's content but in its epistemic commitments, pointing to \kg selection revision rather than content repair;
    \item
        \emph{where the governing schema is inconsistent or \kg content is non-compliant} (an $\DimE$-failure), content repair or schema revision is indicated.
\end{enumerate}
This constitutes an \emph{affordance gap} signal. Unlike abstract
quality metrics~\cite{tartir2005ontoqa,zaveri2016survey}, it
characterises fitness for a specific agentic use and 
identifies the class of remedial actions required.

\subsection{Worked Example}

\begin{table}[t]
    \centering
    \scriptsize
    \caption{The three \kg \AAPs computed against \(\mathcal{S}\)
    ($\DimD$ and $\eta$ reported qualitatively).}\label{tab:example}

    \begin{tabular}{lcccc@{\hspace{1.4em}}l}
        \toprule

& $\DimE\!=\!(\mathcal{L}, \eta)$ & $\DimD$ & $\DimG$ & $\DimR$ & Planner verdict \\
\midrule
\multirow{2}{*}{\textbf{KG\textsubscript{1}}} & 
    \multirow{2}{*}{\textsc{RDFS} (low)} &
    \multirow{2}{*}{low} &
    \multirow{2}{*}{0.5} &
    \multirow{2}{*}{undeclared (Simple)} &
    $\DimG$-failure: no \textsc{TBox} grounding for \\
   &&&&&  $\mathsf{Invited\_speaker}$ \\
\textbf{KG\textsubscript{2}} & \textsc{OWL\,2\,EL} (high) & med  & 1   & \textsc{EL}, no closure
   & $\DimR$-failure: unsafe negative inference \\
\textbf{KG\textsubscript{3}} & \textsc{OWL\,2\,DL} (high) & high & 1   & \textsc{DL}+declared closure
   & all dimensions met; selected \\
\bottomrule
    \end{tabular} 
    \vspace{-0.1in}
\end{table}

To illustrate the framework, imagine that an LLM-orchestrated
assistant is asked by an EKAW conference programme chair to compile an
\emph{emerging-voices} shortlist, comprising researchers who have published
recently on knowledge engineering, yet who have \emph{never} been
invited speakers at a major conference.
The agent's task signature is defined as:
\[
\mathcal{S} = \{\mathsf{Researcher}, \mathsf{Paper}, 
\mathsf{authorOf}, \mathsf{Invited\_speaker}, \mathsf{Conference}, 
\mathsf{givenAt}\}
\]
and its plan requires a \emph{sound} negative
inference on $\mathsf{Invited\_speaker}$. However, under \textsc{OWA}, an empty result means only ``no record of being invited'', which could silently and erroneously admit prior invited speakers into the shortlist.
Three \kgs advertise themselves as candidates:
\begin{itemize}[leftmargin=1cm,labelindent=0.5cm,topsep=0pt]
    \item 
        \textbf{KG\textsubscript{1}} is a legacy \textsc{SPARQL} endpoint exposing \textsc{OAEI}-style IRIs in \textsc{ABox} triples with a minimal \textsc{RDFS} schema declaring only $\mathsf{Researcher}$, $\mathsf{Paper}$, and $\mathsf{authorOf}$ (as most exposed IRIs are undeclared in this schema, its \textsc{ABox} conformance $\eta$ is low), and a \textsc{VoID} description giving only triple counts;
    \item
        \textbf{KG\textsubscript{2}} deploys the \textsc{OWL\,2\,EL} fragment of \textsc{Conference}$^{+}$ (Section~\ref{sec:dimensions}) at an \textsc{OWL\,2\,EL} endpoint, with \textsc{VoID}, \textsc{DCAT}, and \textsc{SPARQL-SD} metadata but no completeness declarations;       
    \item
        \textbf{KG\textsubscript{3}} deploys the full \textsc{OWL\,2\,DL} \textsc{Conference}$^{+}$, carrying an \AAP profile in which the extension of $\mathsf{Invited\_speaker}$ is declared complete via a predicate-level completeness statement~\cite{darari2013completeness}.\footnote{Note that KG$_3$ differs from KG$_2$ along two dimensions
simultaneously: the expressivity regime (\textsc{OWL\,2\,DL} vs
\textsc{OWL\,2\,EL}) and the closure scope (completeness declaration
on $\mathsf{Invited\_speaker}$ vs no such declaration). Isolating the
closure component alone would require a fourth \kg, identical to
KG$_2$ but with the completeness declaration added.}
\end{itemize}
Although the name $\mathsf{Invited\_speaker}$ is definable from a sub-signature
of \textsc{Conference.owl}'s vocabulary~\cite{geleta2016ekaw} and
\emph{appears} lexically in KG\textsubscript{1}, the required \textsc{TBox} axioms
are absent, so the concept and role names $\mathsf{Invited\_speaker}$, $\mathsf{Conference}$,
and $\mathsf{givenAt}$ all lie outside
$\mathcal{V}^{+}(\mathit{KG}_1)$, giving $\DimG = 0.5$ (Table~\ref{tab:grounding}).
KG\textsubscript{2} admits a fully grounded query and correctly
classifies recorded invited speakers.  However, under \textsc{OWA} the agent
cannot conclude $\neg \mathsf{Invited\_speaker}(x)$ from an empty
result: $\DimR$'s closure scope is silent precisely on the predicate
the task depends upon (Table \ref{tab:r-components}).
KG\textsubscript{3}'s completeness declaration over $\mathsf{Invited\_speaker}$ makes the closed-world reading locally sound~\cite{darari2013completeness}. The task's negative inference is sound provided the declaration's stated scope covers the major conferences over which the shortlist criterion quantifies.

\begin{table}[ht]
    \centering
    \scriptsize
    \setlength{\tabcolsep}{11pt}
    \caption{The grounding of each task-signature symbol in the three
    \kgs. A tick indicates the symbol is in $\mathcal{V}^{+}$; a cross
    indicates otherwise.}
    \label{tab:grounding}
    \begin{tabular}{lcccl}
    \toprule
    \textbf{Symbol} & \textbf{KG$_1$} & \textbf{KG$_2$} & \textbf{KG$_3$} & \textbf{Explanation} \\
    \midrule
    $\mathsf{Researcher}$ & \checkmark & \checkmark & \checkmark & Explicit in each schema \\
    $\mathsf{Paper}$ & \checkmark & \checkmark & \checkmark & Explicit in each schema \\
    $\mathsf{authorOf}$ & \checkmark & \checkmark & \checkmark & Explicit in each schema \\
    $\mathsf{Invited\_speaker}$ & $\times$ & \checkmark & \checkmark & Equivalence axiom in \textsc{Conference}$^{+}$ \\
    $\mathsf{Conference}$ & $\times$ & \checkmark & \checkmark & In \textsc{Conference}$^{+}$ TBox \\
    $\mathsf{givenAt}$ & $\times$ & \checkmark & \checkmark & In \textsc{Conference}$^{+}$ TBox \\
    \midrule
    $\DimG$                     & 0.5       & 1          & 1          & \\
    \bottomrule
    \end{tabular}
\end{table}
\begin{table}[t]
    \centering
    \setlength{\tabcolsep}{9pt}
    \scriptsize
    \caption{The $\DimR$ components ($\rho_c$, $\rho_e$, $\rho_s$) of the three \kgs. The $\rho_s$ row shows whether closure is declared on the task-critical predicate ($\mathsf{Invited\_speaker}$).}
    \label{tab:r-components}
    \begin{tabular}{lccc}
    \toprule
        & \textbf{KG$_1$} & \textbf{KG$_2$} & \textbf{KG$_3$} \\
    \midrule
    Consistency status ($\rho_c$) & uncertified & TBox-consistent & Jointly consistent \\
    Declared entailment regime ($\rho_e$) & none        & \textsc{OWL 2 EL} & \textsc{OWL 2 DL} \\
    Closure on $\mathsf{Invited\_speaker}$ ($\rho_s$) & $\times$ & $\times$ & \checkmark \\
    \bottomrule
    \end{tabular}
\end{table}

\medskip
\noindent\emph{Remedial actions.}~The KG\textsubscript{1} verdict
recommends the invocation of a knowledge-producing service that would supply the
missing \textsc{TBox} grounding (e.g., an alignment service projecting
from \textsc{Conference}$^{+}$), whereas the KG\textsubscript{2} verdict
recommends \kg \emph{re-selection} rather than repair, as the
shortfall is with the epistemic commitment, not the content itself (see Table~\ref{tab:r-components} for the specific epistemic-commitment shortfall).
This is the same dimension-to-action mapping as that given in the failure-diagnosis
account (Section~\ref{sec:discussion}): $\DimG$-failures call for vocabulary mediation;
$\DimR$-failures for selection revision; and $\DimE$-failures for content
or schema repair.
The same KG\textsubscript{2} would satisfy the regime requirement $\DimR(\mathit{KG}_2) \succeq \DimR_t^{\min}$ for a task requiring only positive retrieval (e.g., \emph{list current invited speakers}), illustrating the agent- and task-relative character of every dimension.
Large open \kgs such as \textsc{DBpedia} and \textsc{Wikidata}
fall in the low-$\DimD$, \textsc{OWA}/no-closure region of
$\DimR$, making them unsuitable for tasks requiring sound negative
inference without additional completeness declarations. Precise
$\DimG$ computation for such \kgs would require the tooling
identified in Section~\ref{sec:research}.

\subsection{Distinguishing \AAP from Federated Query Source-Selection}

The AAP's dimensions $\DimD$, $\DimG$ and $\DimR$ address concerns
that also arise in a superficially similar setting: federated query
engines that select which \textsc{SPARQL} endpoints to invoke for a
given query. Systems such as \textsc{SPLENDID}~\cite{gorlitz2011splendid},
\textsc{HiBISCuS}~\cite{saleem2014hibiscus} and
\textsc{FedUP}~\cite{aimonier2024fedup} build indices over the
constituent endpoints of a federation and at query planning time, use those indices to prune the set of endpoints to which each triple
pattern will be dispatched. Both settings answer a variant of the question ``which sources are
worth engaging?'', and both rely on metadata pre-published by the
sources. The purposes, however, are fundamentally different, and the
differences illuminate what the \AAP framework contributes
beyond source-selection.

\textsc{SPLENDID}~\cite{gorlitz2011splendid} consumes \textsc{VoID}
descriptions to estimate the cardinality of each triple pattern at
each endpoint, and combines those estimates with join-order search to
compute a cost-optimised query plan that minimises intermediate
results and cross-endpoint communication. The metadata it consumes (e.g., triple counts, predicate partitions, and class partitions) is the same
\textsc{VoID} content that this paper uses to partially inform
$\DimE$ and $\DimD$. 
\textsc{HiBISCuS}~\cite{saleem2014hibiscus} refines the approach by
building a hypergraph over the query's join structure and using
URI-authority patterns to prune the source list.
\textsc{FedUP}~\cite{aimonier2024fedup} targets larger
federations, where per-triple-pattern strategies become prohibitively
expensive, by using compact per-source summaries and an
assignment-based pruning strategy that can prune whole endpoints as opposed to
just individual patterns. All three systems share the same
architectural commitment: source selection is a query-time optimisation
over a fixed federation of individually-queryable endpoints, driven by
metadata about content presence.

The \AAP framework differs from this tradition along three
structural dimensions, each with consequences for what the framework can
express. \emph{First}, the unit of selection is different: federated
engines select sources for a specific query (or sub-query), 
whereas the \AAP selects (or scores) \kgs relative to a task
type, identified via the task signature $\mathcal{S}$ and minimum
epistemic requirement $\DimR_t^{\min}$ before any specific query is
formed. A given task typically corresponds to many queries; conversely, a
single query can serve many tasks. 
Since selection is task-based rather than query-based, all queries
that instantiate a given task select the same \kgs, regardless of
the specific \textsc{SPARQL} phrasing used.
\emph{Second}, the fitness criterion is different: federated engines
score \emph{content availability} (whether triples matching a
pattern exist in the federation), whereas the \AAP scores
\emph{schema and inference adequacy}: whether the source's
vocabulary grounds the task's concepts ($\DimG$), and whether its
entailment regime and closure declarations warrant the required
reasoning ($\DimR$).
A source that would satisfy a federated source-selection index (its
statistics show relevant triples) may still fail the \AAP's
$\DimG$ or $\DimR$ criteria if its schema does not axiomatise the
concepts the task uses, or if its endpoint operates under \textsc{OWA}
without the completeness declarations a negative-inference task
requires. \emph{Third}, the epistemic mode is different: federated
source selection is a \emph{retrieval} discipline, with no explicit
account of what an empty result warrants, whereas the \AAP makes closure assumptions first-class ($\DimR$
component (c), the completeness scope $\rho_s$), so that a task
requiring sound negative inference is correctly ruled in or out
before invocation.

The two settings are complementary rather than competitive. A
deployed agent could use both in sequence, by initially applying
the \AAP to determine which \kgs are task-fit at all, and
then using federated source-selection indices to route specific
queries across the selected \kgs at execution time. From this
perspective, \textsc{VoID}-derived statistics, hypergraph authority
summaries, and assignment-based summaries all count as instances of
the pre-assessable metadata that the \AAP's
$\DimD$ measures. The \AAP extends such indices with two
classes of declaration not carried by current federated indices,
namely task-vocabulary grounding statements that are aligned with a shared
task ontology ($\DimG$-relevant), and predicate-level completeness
or closure declarations ($\DimR$-relevant). A federation whose
sources publish \AAP profiles could therefore support a
two-tier source-selection strategy in which task-level fitness
pruning at planning time is followed by query-level cardinality-based
routing at execution time, with the former eliminating from
consideration the sources the latter would otherwise expend cost
investigating.

\section{Research Agenda}
\label{sec:research}

The framework presented above operates at the vision level.
Realising it as deployable infrastructure requires progress on five
challenges that span formal theory, computational methods, and
the engineering practice needed to make deployment tractable.

\medskip \noindent\textbf{1) AAP Ontology Formalisation:~}
The \AAP vocabulary must be specified as an \textsc{OWL}
ontology enabling formal affordance matching at agent planning time.
Existing standards such as \textsc{VoID}, \textsc{DCAT}, and \textsc{SPARQL-SD} partially constrain $\DimE$, $\DimD$, and the declared-entailment-regime component of $\DimR$, but do not address $\DimG$ or the completeness-scope component of $\DimR$. A new formal annotation vocabulary is therefore required.
A complementary prerequisite is a shared \emph{task ontology} defining
the agent task types against which \AAP scores are computed;
this could initially be bootstrapped from existing Model Context Protocol\footnote{\url{https://modelcontextprotocol.io/docs/getting-started/intro}} (\textsc{MCP}) tool
catalogues and agent capability schemas.

The scope and design of this task ontology is itself a substantive
question. A universal task ontology, of the kind attempted for
upper-level ontologies in the Semantic Web, would be neither
realistic nor necessary. 
The layered scope we take is per-application, but grounded in
existing domain ontologies: a per-application task-signature catalogue
drawn from the concepts and roles of one or more domain ontologies,
rather than a new ontology built from first principles.
The worked example in Section~\ref{sec:vision}
illustrates the pattern: the emerging-voices task's signature
$\mathcal{S}$ is a projection over the vocabulary of
\textsc{Conference}$^{+}$; a citation-recommendation or
co-author-analysis task would draw a different signature over the
same substrate, a medical agent over \textsc{SNOMED-CT}, and so on.
Open questions concern the internal structure of the signature
declarations themselves. A minimal declaration carries only
$\mathcal{S}$; a fuller declaration might also carry the minimum
epistemic requirement $\DimR_t^{\min}$ and, following the \owls
\emph{Profile} pattern of inputs, outputs, preconditions, and
effects~\cite{paolucci2002matching}, epistemic preconditions on the
agent and epistemic contributions to its knowledge state. How to
layer these commitments, how task ontologies extend as agents
encounter new tasks, and how such extensions interact with the
validity of previously-published \AAP profiles are all part
of the same design programme.

\medskip \noindent\textbf{2) Tractability of Task-Relative Grounding:~}
Deciding $\mathcal{V}^{+}$-membership scales with subsumption complexity in the governing fragment, which is tractable in \textsc{OWL\,2\,QL}/\textsc{DL-Lite} and \textsc{OWL\,2\,EL}, expensive in more expressive fragments. The open question is sound approximation, i.e.\ conservative coverage estimates that fail safely, making \AAP matching practical at scale without paying full subsumption cost.

Beyond raw computation, several practical concerns remain open.
Publisher-side caching and reuse strategies could amortise the cost of
$\mathcal{V}^{+}$ construction across the many tasks whose signatures
share substructure, and incremental maintenance of $\mathcal{V}^{+}$
under axiom evolution (a natural extension of the module-extraction
literature) is required if AAPs are to remain valid as the \kg content
changes.
Benchmark suites and empirical protocols for measuring signature-closure
computation on real-world \kgs are also absent, yet are a prerequisite
for calibrating whether approximate matching is sound enough in
practice. The relative strength of approximate versus defeasible matching for
real-world tasks is still to be established.

\medskip \noindent\textbf{3) Compositional Affordance Semantics:~}
When a task requires composition across multiple \kgs, the union coverage of
$\mathcal{V}^{+}$ is necessary but not sufficient, as shared concept
names may carry conflicting axioms and the \kgs' entailment regimes
may not combine coherently (e.g. when predicates fall under mutually incompatible \textsc{LCWA} scopes).
It is unclear how composition across \kgs preserves coherence, or whether mediators preserve meaning when the \kgs' closure assumptions differ (cf.~\cite{daga2026continuum}, Challenges~2 and 3). Existing work on negotiating alignments to limit logical violations~\cite{jimenezruiz2016limiting,payne2014aamas}
offers one approach to preserving coherence.

The same kind of variation can arise within a single \kg. Its
consistency status, entailment regime, or completeness scope may
differ from one named graph to another, or from one sub-endpoint to
another. Accommodating this may require the single-tuple $\DimR$
framing of Section~\ref{sec:dimensions} to be generalised to a set of
context-scoped tuples. Named Graphs~\cite{carroll2005namedgraphs} provide one candidate mechanism
for attaching $\DimR$-tuples to KG-internal scopes. A natural extension would be to develop composition rules for tuples across contexts, together with task-side specifications
of which contexts a task's $\DimR$ requirements apply to.

A parallel question concerns the requesting agent's own store. Once
composition places multiple sources' content into a single agent's
knowledge base, that agent must choose what to retain and under what
warrant. The safest retention policy keeps only assertions expressible in the agent's own
signature, reducing compositional coherence to a local consistency
check, in the formal setting of knowledge base exchange~\cite{arenas2016kbexchange}, at the cost of forgetting anything not
expressible in that signature. Stronger policies exist: retaining the full set of
consequences the transaction entails over that signature requires uniform
interpolation, whose interpolants are not always finitely representable in the
agent's own logic~\cite{lutz2011interpolation}, and retaining the mediating
vocabulary itself reintroduces the compositional hazards this framework was
designed to prevent. The retention policy chosen
determines what a compositional \AAP must record as warrant.

These compositional questions arise regardless of the consumer
architecture that ultimately assembles the composite \AAP:
a discovery intermediary matching against registered profiles, or a
hypermedia agent evaluating affordances resource-by-resource at
navigation time~\cite{Ciortea2019Decade}. The framework remains
neutral over these architectures, exposing the same dimension-level
composition rules to each. A further generalisation extends these
questions beyond \kgs to agents-as-epistemic-resources, opening a
distinct multi-agent research direction in which the four dimensions
apply to agents themselves.

\medskip \noindent\textbf{4) Knowledge Service Specification:~}
For \psms and mediators to be evaluable as planning objects, each must declare its epistemic preconditions and the concepts it produces or bridges. This is \emph{Profile}-level content in the \owls sense:
\iopes were published in the \emph{Profile} precisely so that a requester could assess fitness without invoking the service. What the \AAP provides for \kgs, this item asks for knowledge services.
As current \textsc{MCP} tool descriptions and \textsc{LLM} tool
schemas specify neither, an agent cannot assess a tool's fitness
without invoking it.
This item closes the remediation loop of Section~\ref{sec:cycle}: when $\DimG < 1$, the agent's recourse is to locate a knowledge-producing service supplying the missing grounding, which is only possible if such services carry declarations assessable before invocation.

Ontology-alignment services~\cite{geleta2016ekaw,jimenezruiz2016limiting,agentom2025} are
the most developed instance of this class. Modern alignment systems
already produce correspondences between concept and role names
across heterogeneous vocabularies, but treat their output as a static
artefact rather than as a knowledge service with declared
preconditions. Publishing alignment services against an
\AAP-compatible interface would let an agent evaluate,
before invocation, whether a given alignment supplies the
$\DimG$-relevant coverage its task requires, at what cost, and with
what epistemic preservation guarantees. 
This turns alignment from a one-shot integration step into a planning
object with the same signature vocabulary as the rest of the framework;
where a mediator, a \psm and a \kg are assembled into one access path,
what needs describing is the composed path rather than each component in
isolation, including the requester's own obligations within it.

\medskip \noindent\textbf{5) Engineering Integration and Adoption:~}
A practical deployment requires tooling to assist \kg publishers in
computing \AAP dimension values, and standard protocols for
embedding affordance matching into current agent planning frameworks
without framework-specific adaptation.
Priority targets include \textsc{LLM}-driven frameworks in which
\AAP registry queries serve directly as preconditions on tool
use.

A further concern is the trust layer surrounding \AAP
declarations. Publisher-declared completeness scopes and consistency
statuses can be misstated, either accidentally or adversarially, and the
consequences for the consuming agent's plan can be silent and severe.
Provenance signatures, third-party certification, and mechanisms for
detecting drift between declared and operative regime are all engineering
prerequisites for deployable trust in \AAP-based decisions.
Equally underdeveloped is the empirical protocol: what benchmarks,
task suites, and baselines are appropriate for measuring whether
\AAP-enabled agents plan more reliably than agents relying on
\textsc{VoID} and \textsc{DCAT} alone.
A further challenge is in closing the feedback loop: translating
dimensional failure signals into actionable repair guidance for \kg
maintainers, connecting affordance-based diagnosis to \kg lifecycle
management.

\section{Conclusions}
\label{sec:conclusions}

Two decades ago, the Semantic Web Services community was asked how an
agent can assess a web service resource's fitness before committing to
engage it.
This question applies directly to \kgs, yet remains unanswered by
current standards: \textsc{VoID}, \textsc{DCAT}, and \textsc{SPARQL-SD} describe what
a \kg contains and how its endpoint behaves, but provide no
comprehensive account of what a specific agent can prove from it,
what closure assumptions govern empty results, or whether its task
vocabulary is grounded in the schema.

\looseness=-1
The four \AAP dimensions determine whether a task
is representable ($\DimE$), grounded ($\DimG$), epistemically sound
($\DimR$), and selectable without exploration ($\DimD$),
operationalising the ontological continuum's affordance
structure~\cite{daga2026continuum} at the level of individual
agent \kg-interaction planning.
Together with the companion paper~\cite{daga2026continuum}, which
develops the full ontological continuum and demonstrates its
formalisation, for example through \emph{Formal Concept Analysis}
(\textsc{FCA}), this work forms a coordinated research programme.
The continuum supplies the characterisation space for \kgs at the
community level; the \AAP operationalises the affordance
half of that space at the level of individual agent planning,
publishing pre-computed affordance profiles alongside the signifier
metadata that existing standards already provide and thereby
bringing the signifier/affordance decoupling of
Section~\ref{sec:background} into practical operation.
The framework presented here is a vision-level foundation rather than an operational specification: the compositional-affordance semantics, tractability approximations, and engineering integration identified in the research agenda are still to be developed, and empirical validation from synthetic use cases to real \kgs remains to be done.
Unlike in 2005, mature \textsc{SHACL} tooling, \textsc{OBDA}-aware
triple stores, and \textsc{LLM}-orchestrated agent frameworks make this
substantially more tractable than at the time of \owls and \wsmo. The
research agenda presented here identifies the formal, algorithmic, and deployment challenges that are still to be addressed.

\bibliographystyle{splncs04}
\bibliography{arxiv_V2_checkedref}

\end{document}